\pdfoutput=1

\documentclass[11pt,dvipsnames]{article}

\usepackage[preprint]{acl}

\usepackage{times}
\usepackage{latexsym}
\usepackage{amsfonts}
\usepackage{amsmath}
\usepackage{dsfont}

\usepackage[T1]{fontenc}

\usepackage[utf8]{inputenc}

\usepackage{microtype}

\usepackage{inconsolata}

\usepackage{graphicx}
\usepackage{caption}
\usepackage{subcaption}
\usepackage{tablefootnote}

\usepackage{soul}
\usepackage{booktabs}

\title{Scalable Data Ablation Approximations for Language Models through Modular Training and Merging}

\author{Clara Na\textsuperscript{1,2} \, 
Ian Magnusson\textsuperscript{1,3} \, 
Ananya Harsh Jha\textsuperscript{1,3} \, \\
\textbf{Tom Sherborne\textsuperscript{4}} \, 
\textbf{Emma Strubell\textsuperscript{1,2}} \, 
\textbf{Jesse Dodge\textsuperscript{1}} \,
\textbf{Pradeep Dasigi\textsuperscript{1}} 
\\
  \textsuperscript{1}Allen Institute for AI \,
  \textsuperscript{2}Carnegie Mellon University\,
  \textsuperscript{3}University of Washington\\
  \textsuperscript{4}Cohere\, \\
  \texttt{csna@cs.cmu.edu}}

\begin{document}
\maketitle
\begin{abstract}

Training data compositions for Large Language Models (LLMs) can significantly affect their downstream performance. 
However, a thorough data ablation study exploring large sets of candidate data mixtures is typically prohibitively expensive since the full effect is seen only after training the models; this can lead practitioners to settle for sub-optimal data mixtures. 
We propose an efficient method for approximating data ablations which trains individual models on subsets of a training corpus and reuses them across evaluations of combinations of subsets.
In continued pre-training experiments, we find that, given an arbitrary evaluation set, the perplexity score of a single model trained on a candidate set of data is strongly correlated with perplexity scores of parameter averages of models trained on distinct partitions of that data. 
From this finding, we posit that researchers and practitioners can conduct inexpensive simulations of data ablations by maintaining a pool of models that were each trained on partitions of a large training corpus, and assessing candidate data mixtures by evaluating parameter averages of combinations of these models. 
This approach allows for substantial improvements in amortized training efficiency -- scaling only linearly with respect to new data -- by enabling reuse of previous training computation, opening new avenues for improving model performance through rigorous, incremental data assessment and mixing. 
\end{abstract}

\section{Introduction}
\label{sec:introduction}

\begin{figure}[t]
    \centering
    \includegraphics[width=0.5\textwidth]{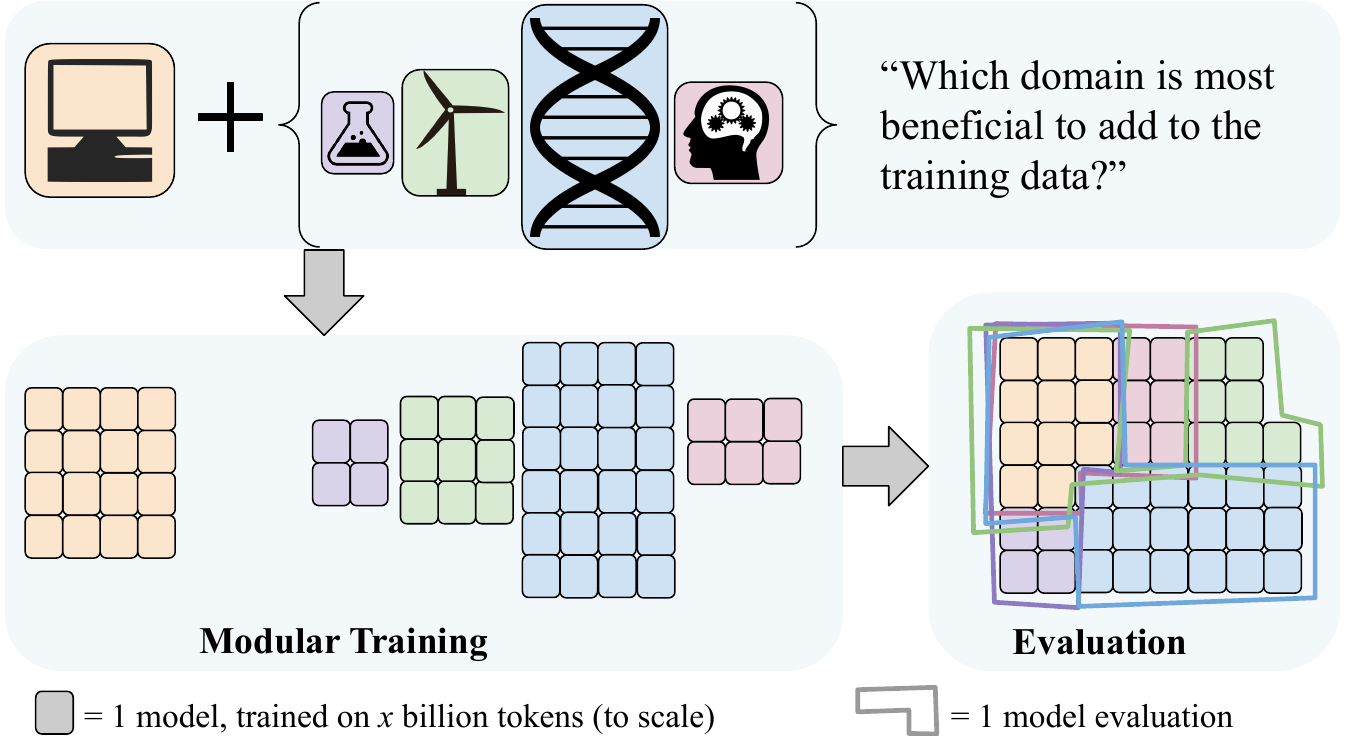}
    \vspace{-1.5em}
    \caption{Given a corpus containing multiple subsets of data, a traditional approach to studying the effects of training on different data mixtures calls for training models on each candidate data mixture. Our strategy \textit{reuses} training on data shared across candidate mixtures, by 1) conducting \textbf{modular training} of many models on equally sized ``base unit'' data partitions and 2) performing \textbf{evaluation} on \textit{parameter averages} of model combinations. We find that this yields useful \textit{proxy} metrics for predicting perplexity scores on arbitrary evaluation domains, such that we can simulate comprehensive data ablation studies at a fraction of the cost. 
    }
    \vspace{-1.5em}
    \label{fig:fig1}
\end{figure}

As Large Language Models (LLMs) and their training corpora have grown in scale, it is increasingly costly not only to \textit{train} an LLM given a fixed recipe, but also to \textit{develop} recipes for training new language models with improved capabilities. Design decisions can span many factors such as model architecture, optimization techniques, and others; one critical aspect of LLM development is \textit{training data composition}.

It is especially important to find a suitable training data recipe as decisions made in the pre-training stage can propagate downstream to domain- and task-specific settings \citep{yadlowsky2023pretraining}. 
Meanwhile, it is common for organizations that develop models to release reports that describe only the optimal configurations that were eventually found (if that) and the capabilities of the final models, with little to no mention of the development process and the suboptimal configurations that were also assessed.

This leaves open questions about the impacts of different pre-training data compositions on language model performance, which is increasingly expensive to explore rigorously as models and corpora grow larger. Though previous studies have empirically tested the effects of different pre-training data compositions \citep{rae2022scaling, longpre2023pretrainers, muennighoff2023scaling} 
or presented strategies for data selection for training data mixtures \citep{Xie2023DoReMiOD, xia2024sheared}, scaling empirical experiments beyond a handful of candidate mixtures tends to be prohibitively expensive. Moreover, there has been limited exploration into \textit{why} certain domains and data mixtures are more suited for certain evaluation domains than others.

We propose an efficient method for approximating language model perplexity performance for a large collection of pre-training data compositions, using only a fraction of the computational cost of training on the full set of data compositions considered. Specifically, we find that a model's perplexity performance on a set of held-out or arbitrary out-of-domain data tends to be strongly correlated to
the perplexity score of a \textit{parameter average} \citep{Izmailov2018AveragingWL, wortsman2022modelsoups} of individual models trained on distinct partitions of the data.

Since in our approach assessing the utility of new data requires training only on the new data, the amount of training needed to assess data mixtures grows only linearly as a function of new data to be assessed, compared to polynomial or combinatorial growth in the naive approach; see \S\ref{subsec:complexity_analysis}.

We run extensive experiments on pre-training language models, primarily in a continued pre-training setting, and evaluating both in- and out-of-domain performance of mixtures of text from different domains. Throughout our experiments (\S\ref{sec:expts}), we vary the compositions of our data mixtures with respect to total size, component data and model size, and component diversity, in order to explore a set of research questions:
\begin{enumerate}
    \item
    Can we predict the perplexity evaluation performance of a model trained on a mixture of data using the performance of models trained on component data \textit{partitions} (\S\ref{subsec:toy_parts})?
    \item Can we simulate data mixing experiments with unevenly sized partitions (\S\ref{subsec:uneven_parts})?
    \item Can we simulate data mixing experiments combining high-level sources (\S\ref{subsec:fos_l1_mix})?
    \item How do our methods apply to larger models (\S\ref{subsec:1b_expts})?
\end{enumerate}

Some of our main findings include:

\begin{enumerate}
    \item On arbitrary evaluation domains of interest (which may be out of distribution), a \textit{parameter average} of individual models, trained in parallel on subsets of a candidate data mixture, can predict perplexity evaluation scores of a model trained on the full data mixture.
    \item If candidate mixtures are comprised of domains that are uneven in size, divide the training corpus they belong to into evenly sized fundamental units to be used as component subsets. To evaluate the candidate data mixtures, consider the performance of \textit{weighted} parameter averages of the component units' models for each mixture.
    \item Expect more reliability at the ``optimal'' end of the performance distribution. 
    It is often easier to find the most favorable data mixtures by our suggested proxy metrics than the least favorable data mixtures. 
\end{enumerate}

Additionally, we release training code, datasets, and models\footnote{\url{https://github.com/clarana/ez-data-ablations} and \href{https://huggingface.co/collections/claran/scalable-data-ablations-6711b7078273ae67b80afc6a}{data and models}}. 
Although the current study is an exploratory investigation limited to 130 million and 1.1 billion parameter models in continued pre-training \citep{gururangan-etal-2020-dont, gupta2023continual} settings with carefully defined domains, we hope our work will prompt further study across additional settings and scales of data and model size. Ultimately, our goal is to enable researchers and practitioners to simulate more principled and thorough data ablation studies as they develop new LLMs and curate training data corpora for them. 


\section{Modular training for data ablations}
\label{sec:modular_training}

We characterize our proposed approach and its asymptotic complexity vs the traditional method for conducting data ablations.

The asymptotic efficiency advantage of our method is comparable to that found comparing a naive recursive implementation of an algorithm with combinatorial complexity, to one that uses memoization. Our method analogously benefits from an effective decomposition of a large training corpus into subsets 
whose corresponding trained models 
are reused repeatedly across many candidate data mixtures (``overlapping subproblems''). Our method also similarly requires additional cache storage for component models, 
relatively inexpensive compared to additional compute. 

\vspace{-3pt}
\subsection{Formalism}
\label{subsec:formalism}

We consider a set of $n$ partitions $\lbrace\mathcal{P}_1, \mathcal{P}_2, ..., \mathcal{P}_n\rbrace$ comprising a corpus $\mathcal{C}$, where we want to understand the effect of 
training a model on different combinations of partitions. 
We define a \textit{data mixture}, $\vec{d}$, as a vector of labels indicating in/exclusion of each partition $i$ in the mixture, $\vec{d}_{i}=\mathds{1}_{\vec{d}}\left(P_{i}\right)$.
We define a \textit{data ablation study} by 1) a \textit{set} of data mixtures $\{\vec{d}\} \subseteq \{0,1\}^n$ that we are interested in evaluating and 2) a set of evaluation domains.

Let $k=||\vec{d}||_1$ denote the number of partitions included in a data mixture, where $k \leq n$. Note that for a fixed $k$, $|\{\vec{d} \in \{0,1\}^n : \sum_{i=1}^{n}{\vec{d_i}}=k\}|$ is equal to $\binom{n}{k}$. The total number of possible $\vec{d}$ for a given $\mathcal{C}$ is the exponential $|\{0,1\}^n|=2^n$.

\subsubsection{Complexity analysis}
\label{subsec:complexity_analysis}
Using the naive approach, the training cost of evaluating a set of training data mixtures $\{\vec{d} \in \{0,1\}^n : \sum_{i=1}^{n}{\vec{d_i}}=k\}$ in full for a fixed $k$ scales with polynomial complexity, $O(n^k)$; the cost of evaluating \textit{all} $\vec{d} \in \{0,1\}^n$ scales exponentially, $O(2^n)$. 

In comparison, our approximation method's cost is $O(n)$ and calls for the caching and reuse of models trained on partitions common across across candidate mixtures $\vec{d}$; thus, training is required only once for each partition. Additional evaluations of data mixtures only require additional training for any previously \textit{unseen} partitions (and caching of the resulting new models).\footnote{Reaping the full computational efficiency benefits of our method requires \textit{caching} of all trained individual models; space complexity is $O(n)$ in our method, whereas the naive method is $O(1)$ and does not require caching trained models.}

\begin{figure}
    \centering
    \includegraphics[width=0.45\textwidth]{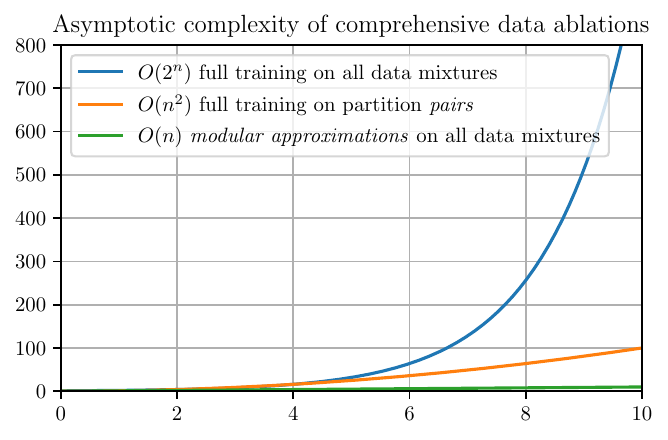}
    \vspace{-1em}
    \caption{Our method (green) 
    has only linear runtime complexity with respect to the number of unique partitions in a corpus. This allows us to simulate \textit{comprehensive} data ablations at extremely low cost compared to naive training on all possible partition combinations.}
    \vspace{-15pt}
\label{fig:asymptotic_complexity}
\end{figure} 

Though in practical settings we may not necessarily want to compare \textit{all} possible data mixtures $\vec{d}$ (and indeed, we do not explicitly verify whether it is possible to simulate an ablation study 
where $\exists~\vec{d}_a, \vec{d}_b \in \{\vec{d}\}$ such that $||\vec{d}_a||_1 \neq ||\vec{d}_b||_1$), the efficiency gains from our approximation method are substantial even when we limit ourselves to studying $\{\vec{d} \in \{0,1\}^n : \sum_{i=1}^{n}{\vec{d_i}}=k\}$ for some fixed $k$ (in Fig.~\ref{fig:asymptotic_complexity}, we have $k=2 \rightarrow O(n^2)$).



\subsection{Method setting and description}

In our work, we explore corpora $\mathcal{C}$ that can be divided into partitions $\mathcal{P}$ using metadata, 
for example along high-level source domains (e.g. academic documents vs source code), topic (e.g. biology vs sociology), or temporality (e.g. news articles by year). 
In conducting a training data ablation study $\{\vec{d}\} \subseteq \{0,1\}^n$, one might wish to evaluate data mixtures' alignments with a handful of specific domains (e.g. certain domains in Paloma; \citet{magnusson2023paloma}), investigate the differential effect of training on particular types of text (e.g. the newest batch of data from a social media website, or scientific documents vs patents), or identify particularly influential or unhelpful subsets of training data for an evaluation domain of interest. 

In practice, partitions of interest may be uneven in size. We find that training and parameter averaging more models on smaller component partitions is favorable to training and parameter averaging fewer models on larger partitions within a larger, unevenly distributed data mixture, which we show in \S\ref{subsec:uneven_parts} and \S\ref{subsec:fos_l1_mix}. Formally, our recommended strategy is as follows:
\vspace{-3pt}

\begin{enumerate}
    \item[1.] Given a corpus $\mathcal{C}$, identify data partitions of interest $\mathcal{P}_1, \mathcal{P}_2, .., \mathcal{P}_n \in \mathcal{C}$, based on ablation studies of interest. These partitions can be of any size. However, assume that within a single ablation study $\{\vec{d}\} \subseteq \{0,1\}^n$, we have some fixed $k = ||\vec{d}_a||_1 = ||\vec{d}_b||_1 \forall~\vec{d}_a, \vec{d}_b \in \{\vec{d}\}$.
    \item[2.] Further \textit{split} (or recombine) partitions $\mathcal{P}_i$ into \textit{equally sized} ``base unit'' partitions $\mathcal{P}_1', \mathcal{P}_2', ..., \mathcal{P}_m' \in \mathcal{C}$. \textit{Train} a copy of a language model on each of the $m$ partitions $\mathcal{P}_j'$, on an equal number of tokens $t$. 
    \item[3.] For each ablation study $\{\vec{d}\} \subseteq \{0,1\}^n$, 
    evaluate data mixtures of $k$ partitions on arbitrary evaluation domains of interest using \textit{parameter averages} of the base component models from Step 2. 
    We posit that researchers and practitioners can use the resulting proxy perplexity scores to identify data mixtures $\vec{d'} \in \{0,1\}^n$ that are better or worse fit to these evaluation domains of interest.
\end{enumerate}

\section{Data}
\label{sec:data}

The training datasets for our main experiments are derived from the Semantic Scholar Open Research Corpus (S2ORC) \citep{Lo2020S2ORCTS} and M2D2 Wikipedia \citep{reid-etal-2022-m2d2}. 
Additionally, for a given model size, we begin by pre-training a model on a general ``seed'' corpus of Gutenberg books and English Wikipedia. All models of the same size used in the current study are initialized from a copy of this seed model before being trained on a combination S2ORC and M2D2 Wikipedia text. 
We evaluate our models using held-out sets from the continued pre-training data, as well as a subset of the validation sets from Paloma \citep{magnusson2023paloma}, an evaluation dataset for assessing language model fit on various domain-specific text. Table~\ref{tab:all_datasets} describes characteristics and sizes of these datasets.

\begin{table}[h]
\footnotesize
    \centering
    \begin{tabular}{p{2.4cm}p{1.8cm}rr}
        Purpose & Data source & \# Part. & Tokens \\ \toprule
        Pretraining (PT) & Wiki+Gutenberg & --- & 10.6B \\ 
        Continued PT, Eval & S2ORC & 128 & 43.9B \\ 
        Continued PT, Eval & M2D2 Wiki & 11 & 2.9B \\ 
        Evaluation only & Paloma\tablefootnote{Paloma \citep{magnusson2023paloma} contains 16 top-level sources and 546 domains, with 123.68 million 
    tokens total; we use a diverse subset of the validation sets, listed in Table~\ref{tab:main_pearson}.} & 9 & 27M \\ \bottomrule 
    \end{tabular}
    \caption{Statistics of datasets. Token counts use the GPT-NeoX-20B tokenizer\tablefootnote{Following \citet{groeneveld2024olmo}, with the addition of special tokens.} \citep{black-etal-2022-gpt}, and include training and held-out sets when applicable. 
    }
    \label{tab:all_datasets}
\end{table}

As ours is an exploratory study, we experiment primarily at 1) smaller scales and 2) with data mixtures of carefully constructed domains \textit{within} datasets of continual pre-training scale. 

Data is partitioned topically and temporally in the case of S2ORC, and topically in M2D2 Wikipedia\footnote{\citet{reid-etal-2022-m2d2} also introduce M2D2 S2ORC, a corpus of academic documents partitioned hierarchically into high- and low-level topical domains. We instead use the larger S2ORC corpus \citep{Lo2020S2ORCTS} in its decontaminated, Dolma-like \citep{dolma} format, which retains document boundaries not easily recoverable from the M2D2 format}. For M2D2 Wikipedia, we use the L1 (top-level) domains as originally defined by \citet{reid-etal-2022-m2d2} with a median of 282 million tokens each. As S2ORC is not pre-partitioned, we construct 22 partitions $\mathcal{P}$ for S2ORC based on metadata for field of study (FoS), and 128 ``base unit'' partitions $\mathcal{P'}$ of roughly equal size, around 287 million tokens each, based on FoS and year. 

Some documents are tagged with multiple fields of study (see Table~\ref{tab:s2orc_multiple_fos} in Appendix~\ref{sec:appendix_s2orc}) in which case we use the first FoS listed. We treat the $1\%$ of documents without an explicit tagging as their own FoS, 
indicated by ``NA''. Documents' publication years range from 1970 to 2022. All partitions are associated with one or more publication years and a single field of study (with ``NA'' treated as its own FoS), with two exceptions: Geography was merged into Environmental Science, and Law was merged into Political Science, due to their high co-occurrence and relatively small standalone size. 

Table~\ref{tab:s2orc_sizes} in Appendix~\ref{sec:appendix_s2orc} shows partition counts for each field of study: fields of study with larger partition counts are larger than fields of study with smaller partition counts. Humanities fields are overall less heavily represented in the corpus; many have only a single partition (containing all documents from 1970 through 2022), while some of the largest partitions contain over 1 billion tokens from papers published in a single year within a field of study. In practice, in \S\ref{sec:expts}, we \textit{upsample} smaller partitions when training such that data mixtures from multiple partitions within S2ORC include roughly equal representation from each partition. 

All seed and continued pre-training data was \textit{decontaminated} against Paloma evaluation data, and all continued pre-training data was decontaminated against its respective evaluation splits. Following \citet{dolma}'s standard for decontamination on Paloma \citep{magnusson2023paloma}, a Bloom filter was used to exclude all documents containing least one ``contaminated'' paragraph (at least 13 tokens long and found in the evaluation set). 
In general, we use S2ORC and M2D2 Wikipedia as examples of datasets at continued pre-training scale that can be partitioned along natural inherent groupings in their distributions, but we encourage future work exploring different data.

\section{Methods}
\label{sec:methods}
For our base experiments, we train small (130m parameter) decoder-only models with a PaLM-like \citep{Chowdhery2022PaLMSL} 
architecture, on independent partitions of the Semantic Scholar Open Research Corpus (S2ORC) \citep{Lo2020S2ORCTS} and M2D2 Wikipedia \citep{reid-etal-2022-m2d2}, defined as discussed in \S\ref{sec:data}. In a subset of our experiments, we involve larger (1.1b parameter) models. Models share an initial optimization trajectory of 
~42 billion tokens of ``seed'' pre-training on a general corpus of 10 billion tokens consisting of Gutenberg books and English Wikipedia. Each model is then trained on a specific partition within S2ORC or M2D2 Wikipedia for some multiple of 1 billion tokens, depending on the experimental setup. 
We evaluate models using per-token perplexity scores on domain-specific validation sets as well as out-of-domain sets from Paloma \citep{magnusson2023paloma}; all perplexity scores reported in this study are on held-out sets. We use the GPT-NeoX-20B tokenizer \citep{black-etal-2022-gpt} as it is used by \citet{groeneveld2024olmo} with the addition of 3 special tokens for masking personally identifiable information\footnote{\href{https://huggingface.co/allenai/gpt-neox-olmo-dolma-v1_5/tree/main}{gpt-neox-olmo-dolma-v1\_5}}.

See Appendix~\ref{sec:appendix_hyperparams} for more detailed reporting on training settings, hyperparameters, and hardware.

\section{Experiments}
\label{sec:expts}

In our experiments, we have:
{\footnotesize
\begin{align*}
    \mathcal{C} &= \bigcup_{f=1}^{22}{\mathcal{P}_{S2ORC_{f}}} \cup \bigcup_{g=1}^{11}{\mathcal{P}_{Wiki_{g}}} \\
    &= \bigcup_{i=1}^{128}{\mathcal{P}'_{S2ORC_{i}}} \cup \bigcup_{g=1}^{11}{\mathcal{P}'_{Wiki_{g}}}
\end{align*}
}%
Partitions $\mathcal{P}' \in \mathcal{C}$ are disjoint, but 
we have no evidence to suggest 
that this is a strict requirement.\footnote{In fact, our $\mathcal{P}' \in \mathcal{C}$ almost certainly have \textit{distributional} overlap, though very significant overlap may reduce our method's effectiveness.} Total number of partitions is $n=22+11=33$ in most of our experiments, except in \S\ref{subsec:toy_parts} where we sample partitions directly from $\{\mathcal{P'} \in \mathcal{C}\}$ and so $n=128+11=139$. The set of evaluation domains in each experiment includes 1) in-domain datasets from each data mixture $\vec{d}$'s respective combined held-out dataset and 2) nine domains from Paloma as OOD or domain-shifted data.



\subsubsection{Figure information}
\label{subsec:figure_info}
In all experiments, for each evaluation dataset, we plot the perplexity evaluation performance of the model trained sequentially\footnote{We describe training as ``sequential'' to distinguish from the modular training on individual partitions of data that can be performed in parallel, \textit{not} to refer to any particular curricular ordering of the training data.} on the entire data mixture (\textsc{seq}) to various \textit{proxy} perplexity evaluations: each point lies along the same $y=a$ as at least one other point associated with the same data mixture (and a different proxy signal). We compare \textsc{seq} model scores to the average of the perplexity scores of the individual (\textsc{ind}) models trained on the components of the data mixture, as well as to the perplexity scores of (\textsc{merged}) parameter averages of base component models. These \textsc{ind} models' average scores are written as ``mean \textsc{ind}'' in the figures. ``\textsc{ind} ID'' scores are similar but differ in that they include only in-domain held-out evaluations in the average).


In \S\ref{subsec:uneven_parts}, we discuss effective vs ineffective partitioning and weighting for data mixtures with larger or uneven domains. In particular, we compare parameter averages of $k=||\vec{d}||_1=2$ unevenly trained component models with ``macro''- and ``micro''-(parameter) averages of many more, evenly trained ``base unit'' models trained on individual base components $\mathcal{P'}$. In a macro-\textsc{merged} model, the $k$ high-level component models are themselves parameter averages of unequal numbers of equally trained base unit models; a micro-\textsc{merged} model is the direct uniform parameter average of the equally trained base unit models. We find that macro-\textsc{merged} model performance tends to be the most reliable proxy metric in settings where the partitions are sampled from the same distribution of sizes, while micro-\textsc{merged} model performance is more reliable in settings with an expected skew. 

Overall, we find that \textsc{seq} scores on arbitrary evaluation domains (which may be OOD or domain-shifted with respect to the training data) are strongly correlated with the perplexity scores of appropriately composed \textsc{merged} models, though there are settings where mean scores of \textsc{ind} models are also strongly correlated.

\subsection{Base partition studies}
\label{subsec:toy_parts}

\begin{figure*}[h!]
    \centering
    \begin{subfigure}{\textwidth}
        \includegraphics[width=0.345\textwidth]{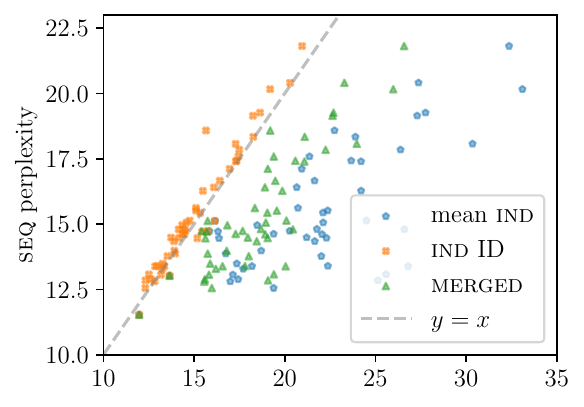}
        \hfill
        \includegraphics[width=0.32\textwidth]{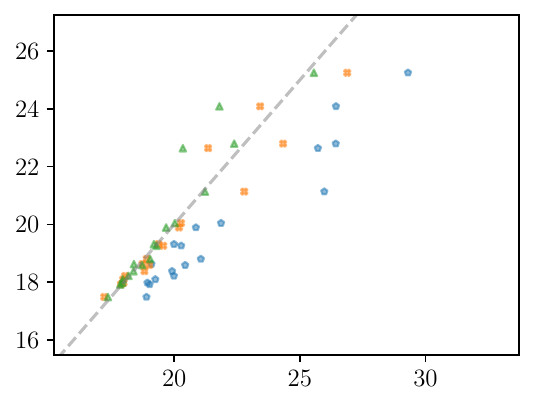}
        \hfill
        \includegraphics[width=0.32\textwidth]{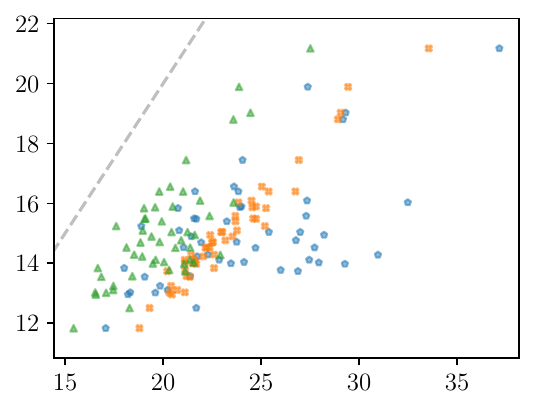}
    \end{subfigure}
    \\
    \vfill
    \begin{subfigure}{\textwidth}
        \includegraphics[width=0.34\textwidth]{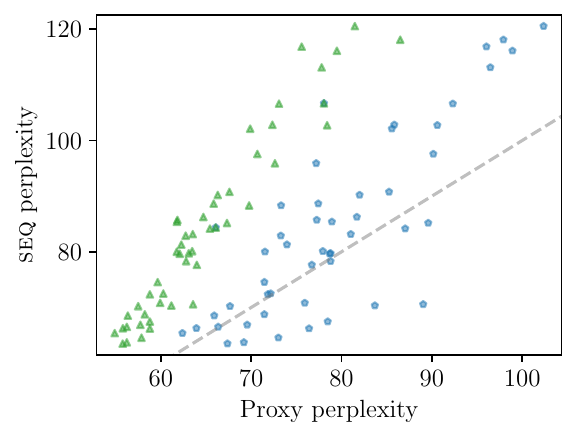}
        \hfill
        \includegraphics[width=0.315\textwidth]{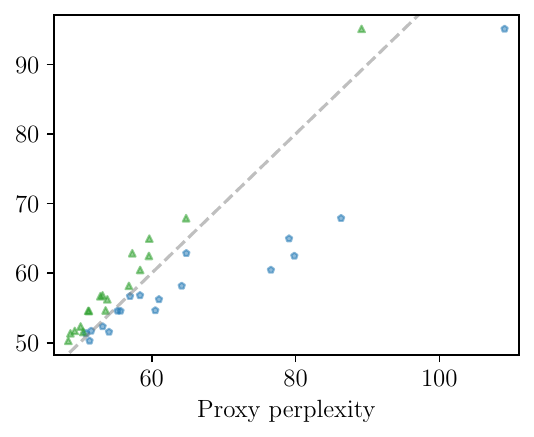}
        \hfill
        \includegraphics[width=0.32\textwidth]{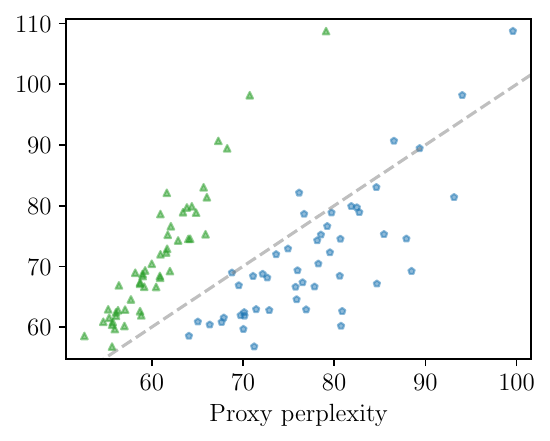}
    \end{subfigure}
    \vspace{-20pt}
    \caption{\textsc{seq}(uentially trained), \textsc{merged}, and \textsc{ind}(ividual) models trained on random pairs of S2ORC base partitions (\textbf{left}), pairs of M2D2 Wikipedia partitions (\textbf{middle}) and triples of S2ORC base partitions (\textbf{right}), evaluated on respective held-out sets of the same (\textbf{top}) and on nine subsets of Paloma (\textbf{bottom}). Each point lies along the same $y-$value for \textsc{seq} score as the other proxy evaluation score(s) for the same data mixture. For held-out sets of the training data, the \textsc{seq} performance on the combined dataset correlates most strongly with the average of the component models' respective in-domain evaluations (``\textsc{ind} ID''), compared to \textsc{merged} model performance or mean \textsc{ind} scores. However, on OOD Paloma data, the Pearson's correlation is highest between \textsc{seq} and \textsc{merged} models. See Table~\ref{tab:pearson_abridged} for correlation values and \S\ref{subsec:figure_info} for definitions of terms.} 
    \label{fig:toy_3expt}
    \vspace{-6pt}
\end{figure*}

We begin with a controlled continued pre-training setting, in which which we select training data mixtures comprised of \textit{pairs} ($k=2$) or \textit{triplets} ($k=3$) of equally sized data partitions $\mathcal{P'}$ (not $\mathcal{P}$ as we do in other experiments) from the same high-level data source domain. We replicate the $k=2$ experiments in both S2ORC ($|\{\vec{d'}\}|=50$) and M2D2 Wikipedia ($|\{\vec{d'}\}|=20$), where data mixtures $\vec{d'}$ are sampled uniformly and without replacement from 
the 128 topical-temporal partitions in S2ORC we have defined in \S\ref{sec:data} (and none from Wikipedia; $\{\vec{d'} \subseteq \{0,1\} : ||\vec{d'}_{S2ORC}||_1=2, ||\vec{d'}_{Wiki}||_1=0\}$), and from the 11 L1 topical partitions as originally defined by \citet{reid-etal-2022-m2d2} ($\{\vec{d'} \subseteq \{0,1\} : ||\vec{d'}_{S2ORC}||_1=0, ||\vec{d'}_{Wiki}||_1=2\}$). For the $k=3$ experiment, we sample 50 mixtures from $\{\vec{d'} \subseteq \{0,1\} : ||\vec{d'}_{S2ORC}||_1=3, ||\vec{d'}_{Wiki}||_1=0\}$.

Starting with the same seed model initialization in all cases, we continue pre-training on these mixtures for 1 billion tokens for each partition. 

Figure~\ref{fig:toy_3expt} displays the strong relationships we find between \textsc{seq} models' performance and our proxy metrics of choice for all three of these settings, and Table~\ref{tab:pearson_abridged} reports the correlation scores. The data mixtures vary in component domain similarity -- for example, in the S2ORC settings, some pairs consist of older and newer partitions from the same FoS, while other pairs differ in both recency and topic. Although greater intra-mixture similarity is related to lower perplexity scores on the held-out sets of training data mixtures, the strong correlation between \textsc{seq} and proxy model perplexities is seen in both low and high perplexity scores. 
\begin{table}[h]
    \centering
    \scriptsize
    \begin{tabular}{llccc}
         $\mathcal{P'}$ Expt & Eval & \textsc{ind} ID & mean \textsc{ind} & \textsc{merged} \\
         \toprule
         S2ORC & ID & \textbf{0.968} & 0.711 & 0.846 \\
            ~~$k=2$ & Paloma & - & 0.819 & \textbf{0.961} \\
         \midrule
         Wiki & ID & 0.962 & \textbf{0.966} & 0.948 \\
            ~~$k=2$ & Paloma & - & 0.942 & \textbf{0.993} \\
         \midrule
         S2ORC & ID & \textbf{0.977} & 0.584 & 0.77 \\
             ~~$k=3$ & Paloma & - & 0.799 & \textbf{0.951} \\
         \bottomrule
    \end{tabular}
    \caption{Pearson's correlation scores between \textsc{seq} model perplexity scores and proxy perplexity scores for experiments in \S\ref{subsec:toy_parts}.}
    \label{tab:pearson_abridged}
\end{table}

In the following sections, we present studies exploring larger data mixture sizes (\S\ref{subsec:uneven_parts}), lower intra-mix similarity (\S\ref{subsec:fos_l1_mix}), and larger model sizes (\S\ref{subsec:1b_expts})

\subsection{Varying data mixture partition sizes} 
\label{subsec:uneven_parts}

\begin{table}[h]
\scriptsize
    \centering
    \begin{tabular}{lcc|ccc}
        & \multicolumn{2}{c}{\underline{Average \textsc{ind} scores}} &  \multicolumn{3}{c}{\underline{~~~~~~~~\textsc{merged} scores~~~~~~~~}} \\
        & \textsc{merged} &	\textsc{\textsc{seq}} &	\textsc{seq} & macro- &	micro- \\
        \toprule
        $\mathcal{P}_1$ & 0.844 & 0.826 & 0.763 & 0.929 & 0.937\\
        $\mathcal{P}_2$ & 0.909 & 0.930 & 0.914 & 0.959 & 0.946\\
        $\mathcal{P}_1+\mathcal{P}_2$ & 0.856 & 0.844 & 0.755 & 0.944 & 0.938\\
        \midrule
        M2D2 S2 & 0.869 & 0.908 & 0.608 & 0.894 & \textbf{0.918}\\
        M2D2 Wiki  & 0.905 & 0.886 & 0.822 & \textbf{0.966} & 0.858\\
        Wiki-103 & 0.927 & 0.885 & 0.783 & \textbf{0.983} & 0.867\\
        PTB & 0.880 & 0.835 & 0.694 & \textbf{0.929} & 0.773\\
        4chan & 0.874 & 0.883 & 0.866 & \textbf{0.905} & 0.785\\
        c4-en  & 0.905 & 0.863 & 0.798 & \textbf{0.985} & 0.849\\
        mc4-en  & 0.844 & 0.836 & 0.770 & \textbf{0.969} & 0.829 \\
        RedPajama  & 0.790 & 0.902 & 0.838 & \textbf{0.930} & 0.852\\
        Manosphere  & 0.917 & 0.919 & 0.895 & \textbf{0.976} & 0.867\\ \midrule
        Avg (macro)  & 0.910 & 0.882 & 0.790 & \textbf{0.984} & 0.848\\ \bottomrule
    \end{tabular}
    \caption{Pearson correlation scores for our experiment mixing unevenly sized partitions, described in \S\ref{subsec:uneven_parts}. We report correlation scores between \textsc{seq} models' perplexity evaluation scores vs candidate proxy evaluation scores, on various held-out and OOD sets. The first three rows are evaluations on held-out sets from respective data mixtures $\vec{d}$ and have \textsc{ind} ID scores > 0.98. The two labels in the first row distinguish between averages of ``\textsc{ind}'' perplexity evaluations (which may themselves be of \textsc{merged} parameter averages or individual \textsc{seq} models), and evaluations of ``\textsc{merged}'' parameter averages of models (whose components may be \textsc{seq} models, \textsc{merged} models, or base component models).}
    \vspace{-5pt}
    \label{tab:main_pearson}
\end{table}

\begin{figure}[h]
    \centering
    \includegraphics[width=0.45\textwidth]{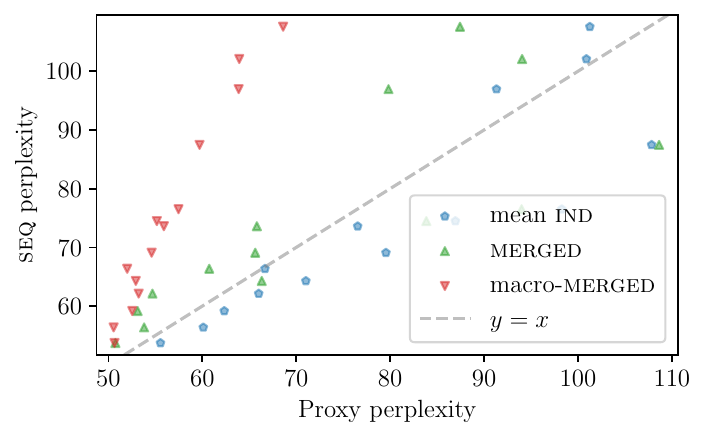}
    \caption{\textsc{seq} vs. proxy macro-averaged Paloma perplexity scores of models trained on differently sized partitions $\mathcal{P}$. We see that the macro-\textsc{merged} scores are the most reliable metric (Pearson's correlation 0.984).} 
    \label{fig:s2orc_2fos}
\end{figure}

We analyze a setting with larger, unevenly sized $\mathcal{P}$ with $k=2$. We verify that models trained for 1 billion tokens each on base unit partitions $\mathcal{P'} \in \mathcal{P}$ exhibit evaluation scores in their parameter averaged forms predictive of \textsc{seq} scores $\forall~\mathcal{P} \in \mathcal{C}$ (see Appendix~\ref{sec:appendix_s2orc_fos}). We then move to sampling from the $n=33$ higher-level partitions $\mathcal{P}$, starting first with pairs ($k=2$) of fields of study from S2ORC: we sample 14 data mixtures from $\{\vec{d} \subseteq \{0,1\} : ||\vec{d}_{S2ORC}||_1=2, ||\vec{d}_{Wiki}||_1=0\}$.

Different from before, we compare \textit{multiple} types of parameter averaged models as candidate proxies for the \textsc{seq} model in each data mixture. In one variation (\textsc{merged}), we evaluate the parameter average of the two models that were each trained for potentially \textit{unequal} optimization trajectory lengths on \textit{unequal} amounts of data. Each component model in this case was trained on data for an entire FoS for a \textit{multiple} of 1 billion tokens, where the multiplier was $|\mathcal{P}|$, the number of base units $\mathcal{P'}$ comprising the field of study partition $\mathcal{P}$.

We describe the other two types of parameter averages as \textit{macro}-\textsc{merged} and \textit{micro}-\textsc{merged} models, where the fundamental component models were each trained for only 1 billion tokens on a single base unit temporal partition $\mathcal{P'}$ belonging to the field of study $\mathcal{P}$. Each macro-\textsc{merged} model in this experiment is created by 1) forming the uniform parameter average using base models trained on each $\mathcal{P'} \in \mathcal{P}$ from $\vec{d}$, and then 2) forming the uniform parameter average of the $k=2$ resulting parameter averaged models. Each micro-\textsc{merged} model is a uniform parameter average of all individual base models trained on each $\mathcal{P'}$ in both $\mathcal{P}$ in the data mixture. Each \textsc{seq} model in this study was trained for $(|\mathcal{P}_1| + |\mathcal{P}_2|) * 1$ billion tokens.

Table~\ref{tab:main_pearson} shows Pearson correlation scores between \textsc{seq} and proxy metrics for the experiment depicted in Figure~\ref{fig:s2orc_2fos}. 
Macro-\textsc{merged} model evaluations are the most useful proxy for \textsc{seq} perplexity evaluation scores in this setting, where, notably, though the data mixtures can contain partitions of uneven size, the \textit{expected values} are equal because they are sampled from the same distribution.

\subsection{Mixed sources}
\label{subsec:fos_l1_mix}

\begin{figure}[h]
    \vspace{-1em}
    \includegraphics[width=0.45\textwidth]{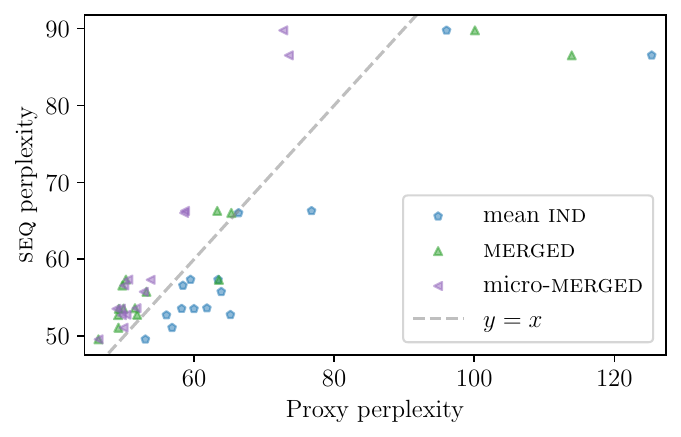}
    \vspace{-1em}
    \caption{\textsc{seq} vs. proxy macro-averaged Paloma perplexity scores of models trained on data mixtures containing data from multiple sources. As in \S\ref{subsec:uneven_parts}, it is beneficial to merge models trained on similar amounts of data, but here, \textit{micro-}\textsc{merged} models are more useful -- 0.989 corr. with \textsc{seq}, vs. 0.963 (\textsc{merge} of the \textsc{seq} models for each $\mathcal{P}$), 0.916 (mean \textsc{ind} scores), 0.807 (macro-\textsc{merged}), or 0.759 (mean \textsc{merged} scores).}
    \label{fig:fos_l1_combo}
\end{figure}

We present a setting where data mixtures have multiple high-level source domains in their partitions, such that the documents in a data mixture are not necessarily similar in format. We again have $k=2$ and $n=33$, and we sample 15 data mixtures from $\{\vec{d} \subseteq \{0,1\} : ||\vec{d}_{S2ORC}||_1=1, ||\vec{d}_{Wiki}||_1=1\}$. We assign training durations to our component and \textsc{seq} models according to the rules of \S\ref{subsec:uneven_parts}, but as $\mathcal{P}_2$ is always a Wikipedia L1 domain, we have $|\mathcal{P}_2|=1~\forall~\vec{d}$ and therefore \textit{more uneven} data mixtures. In Figure~\ref{fig:fos_l1_combo}, we show that when when data mixtures combine these high-level domains, the micro-\textsc{merged} models for each data mixture are highly correlated with \textsc{seq} models in their perplexity scores on arbitrary evaluation domains.

\subsubsection{Fixing one data source}
\label{subsec:fixwiki}

In a variation of the previous experiment, we fix the specific L1 domain (we choose $\mathcal{P'}_{Tech}$: ``Technology and applied sciences'' $\subset \mathcal{P}_{Wiki}$) across all experiments: we use all 22 data mixtures from $\{\vec{d} \subseteq \{0,1\} : ||\vec{d}_{S2ORC}||_1=1, \mathcal{P'}_{Tech} \in \vec{d}\}$.

We argue that our approximation method can be applied in a practical setting where, for example, we wish to find auxiliary data that is beneficial to some existing training data, with respect to some evaluation domain(s); training on a performant data mixture selected by proxy metric yields a performant \textsc{seq} model. Results are discussed with Figure~\ref{fig:fos_fixL1_combo} in Appendix~\ref{sec:appendix_fixwiki}. 

\subsection{Applicability to larger model scales}
\label{subsec:1b_expts}

\begin{figure}
    \centering
    \includegraphics[width=\linewidth]{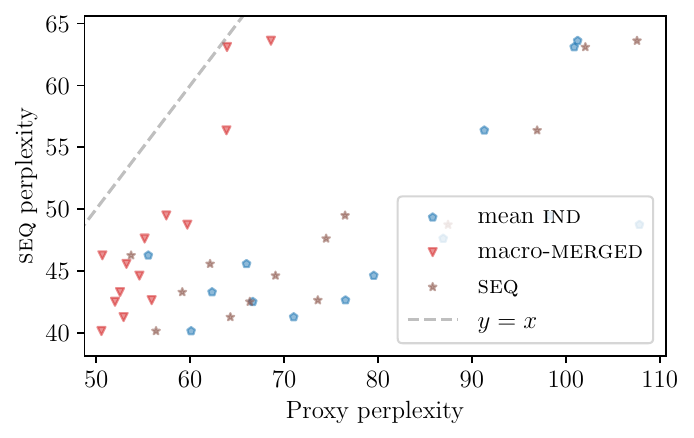}
    \vspace{-2em}
    \caption{Macro-averaged Paloma perplexity scores of 1.1b \textsc{seq} models vs. proxy scores of 130m models trained on differently sized partitions $\mathcal{P}$. We again see that the macro-\textsc{merged} scores are the most reliable metric, with a Pearson's correlation of 0.926. Notably, macro-\textsc{merged} scores of the 130m models are on average more predictive of the 1.1b models' \textsc{seq} scores than the 130m models' own \textsc{seq} scores.}
    \label{fig:1b_pred_fos}
    \vspace{-1em}
\end{figure}

We also experiment with larger (1.1 billion parameter) models and present evidence that our approximation methods are useful beyond the smaller model scale we primarily experiment with, in both the sense that 1) proxy evaluations of 1.1b parameter models can be used to select performant data mixtures, and that 2) proxy evaluations of 130m parameter models can be used to select performant data mixtures for 1.1b parameter models. The latter finding has significance as an \textit{orthogonally} beneficial feature of our proposed approximation method,\footnote{In fact, \citet{ye2024data} shows that smaller (\textsc{seq}) models can predict larger (\textsc{seq}) model performance on a data mixture. While we focus on efficiency gains from reusing training performed on decomposed subparts of an arbitrary data mixture, we show that these efficiency gains are \textit{compatible}.} along with the asymptotic complexity benefit. We highlight the latter case by adapting the previously studied experimental setup with $k=2$ larger, unevenly sized $\mathcal{P}$ from S2ORC (depicted in Figure~\ref{fig:s2orc_2fos} and Table~\ref{tab:main_pearson} from \S\ref{subsec:uneven_parts}) to compare the same proxy model metrics with 1.1b \textsc{seq} model scores on the same candidate data mixtures. The results of this experiment are shown in Figure~\ref{fig:1b_pred_fos} and Table~\ref{tab:1b_pred_pearson}.

\begin{table}[h]
\scriptsize
    \centering
    \begin{tabular}{p{1.15cm}cc|ccc|c}
        & \multicolumn{2}{c}{\underline{Avg \textsc{ind} scores}} & \multicolumn{3}{c}{\underline{~~~~\textsc{merged} scores~~~~}} & \multicolumn{1}{c}{\underline{\textsc{ind}}} \\
        & \textsc{merg.} &	\textsc{\textsc{seq}} &	\textsc{seq} & macro & micro & \textsc{seq}\\
        \toprule
        $\mathcal{P}_1$&0.785&0.751&0.695&0.898&0.872&0.983\\
        $\mathcal{P}_2$&0.896&0.915&0.882&0.952&0.908&0.990\\
        $\mathcal{P}_1+\mathcal{P}_2$&0.802&0.785&0.701&0.911&0.891&0.986\\
        \midrule
        M2D2 S2&0.720&0.829&0.467&0.716&0.740&0.909\\
        M2D2 Wiki&0.864&0.855&0.810&0.945&0.861&0.972\\
        Wiki-103&0.881&0.853&0.755&0.954&0.851&0.977\\
        PTB&0.709&0.636&0.475&0.780&0.576&0.791\\
        4chan&0.778&0.613&0.631&0.914&0.760&0.849\\
        c4-en&0.800&0.720&0.647&0.935&0.789&0.930\\
        mc4-en&0.660&0.487&0.429&0.793&0.650&0.739\\
        RedPajama&0.787&0.808&0.738&0.938&0.862&0.926\\
        Manosphere&0.854&0.832&0.816&0.948&0.833&0.967\\
        \midrule
        Avg&0.810&0.729&0.635&0.926&0.772&0.907\\
        \bottomrule
    \end{tabular}
    \caption{For the experiment described in \S\ref{subsec:1b_expts}, Pearson correlation scores between 1.1b \textsc{seq} models' perplexity evaluation scores vs candidate proxy evaluation scores of 130m models, on various held-out and OOD sets. Here, the \textsc{seq} scores of 130m \textsc{seq} models are included for comparison.}
    \label{tab:1b_pred_pearson}
\end{table}

In Appendix~\ref{sec:appendix_1b}, we discuss additional experiments with 1.1b models, where we study all $\{\vec{d} \subseteq \{0,1\} : ||\vec{d}_{S2ORC}||_1=2, ||\vec{d}_{Wiki}||_1=0, |\mathcal{P}| = 1 ~\forall~ \mathcal{P} \in \vec{d}\}$.

\subsection{Efficiency analysis}
\label{subsec:empirical_efficiency}

In general, our runtime complexity analysis from \S\ref{subsec:complexity_analysis} translates reliably into empirical efficiency gains. We provide concrete examples of computational cost savings from our experiments. Over the 14 candidate mixtures sampled for the experiment in \S\ref{subsec:uneven_parts}, we formed the macro-\textsc{merged} models using 104 total ``base unit'' 130m models that were each trained for 1 billion tokens in around 5 hours on a single A100 GPU, amounting to 520 total GPU hours. The \textsc{seq} models for the candidate data mixtures required 163 billion tokens of training in total. Then, a lower bound estimate for the number of GPU hours used to train the corresponding 14 data mixtures is $163*5=815$ – in reality, the sum was higher due to communication overhead from training \textsc{seq} models on multiple GPUs.

Notably, by our proposed paradigm, we can evaluate the remaining 157 
pairs of fields of study with \textit{no additional training}. We can evaluate all 231 
possible FoS-pair data mixtures with only 120 
additional GPU hours (to include the 3 fields of study that were not represented in the 14 candidate mixtures we sampled), which brings the total GPU hours needed to only 640. 
In contrast, naively training \textsc{seq} models for all possible candidate mixtures would have cost us 2688 
GPU hours total.

That being said, \textit{illustrating} that there exists a strong correlation trend is computationally expensive; our bottleneck is training the very \textsc{seq} models that our method allows us to avoid, or at the very least reduce. Thus, our empirical experiments involve only a fraction of the total search space for their respective settings.

Our method's efficiency advantage is greater and more straightforward to calculate for the experiments in \S\ref{subsec:toy_parts}, since we sample partitions directly from ``base components'' $\mathcal{P'} \in \mathcal{C}$. For the S2ORC-only $k=2$ and $k=3$ experiments, we can use the same 128 base component models from \S\ref{subsec:uneven_parts} (no additional training cost). Naively training models on all possible combinations of data would result in $\binom{128}{2}*2=$16,256 billion tokens of training over 8128 candidate mixtures, for 81,280 GPU hours for $k=2$. For $k=3$, we would have $\binom{12,8}{3}*3=$1,024,128 billion tokens over 341,376 candidate mixtures, for 5,120,640 GPU hours total.

\section{Related Work}
\label{sec:related_work}

\paragraph{Pretraining mixtures} Existing literature includes many open corpora that are presented as viable training data mixtures (or components of them) for pre-training LLMs \citep{dodge-etal-2021-documenting, Gao2020ThePA, together2023redpajama, penedo2023falconrefinedweb, dolma}. The individual components of these mixtures and their relative sizes are typically chosen based on some intrinsic measure of data quality as it is often prohibitively expensive to perform thorough data ablations to create mixtures optimizing for downstream performance.

\paragraph{Efficient data selection} Given that data ablations on large language models is expensive, one class of approaches relies on approximating them on smaller models. Relevant work studies scaling laws for model parameters vs training tokens \citep{hoffmann2022training, biderman2023pythia}, empirical effects of including or excluding different sources of data \citep{longpre2023pretrainers}, and the effects of training over multiple epochs vs new training tokens \citep{muennighoff2023scaling}. Previous work has also explored improving domain-specific fit via continued pre-training \citep{gururangan-etal-2020-dont}, predicting domain fit using lexical features \citep{reid-etal-2022-m2d2}, or improving general test-time adaptation via dynamic data selection, either by distributionally robust optimization with a small proxy model \citep{oren-etal-2019-distributionally, Xie2023DoReMiOD} or online using a multi-armed bandit approach \citep{albalak2023efficient}. Additional previous works aim to adapt to known downstream tasks via data selection, including at the individual example level \citep{wang2020optimizing} or even by explicitly fine-tuning models on many tasks \citep{aghajanyan-etal-2021-muppet}. Notable recent and concurrent work proposes scalable influence functions, traditionally used at only very small scales, as a method for selecting better training data \citep{choe2024dataworthgptllmscale, yu2024matesmodelawaredataselection}. Another concurrent preliminary work \citep{thrush2024improvingpretrainingdatausing} proposes a training-free method of selecting data to improve performance on downstream tasks. See \citet{albalak2024survey} for a survey on data selection methods. We note that our strategies may be compatible with many of the existing methods for dataset selection, potentially leading to cumulative efficiency improvements -- in particular, we show that our proxy metrics are compatible with \citet{ye2024data}'s strategy of predicting larger models' performance on a data mixture using a smaller, proxy model's performance.

\paragraph{Model averaging} Merging models via weight-space averaging is more commonly done in the finetuning stages of language model training~\citep{Izmailov2018AveragingWL, wortsman2022modelsoups}, typically for improving robustness and mitigating cross-task interference. When used for pretraining language models~\citep{Li2022BranchTrainMergeEP, chronopoulou-etal-2023-adaptersoup}, the goal is to efficiently adapt to new domains at inference time. To the best of our knowledge, ours is the only work that studies model merging for efficient data ablations.

\paragraph{Linear mode connectivity} Relevant to our work, the mode connectivity perspective argues models share behavior when linearly connected on the loss surface \citep{pmlr-v119-frankle20a,juneja2023linear, NEURIPS2020_0607f4c7}. Notably, \citet{NEURIPS2018_be3087e7} previously demonstrated that model ensembling is most effective when combining models on the equivariant loss curve on the loss surface. This applies to our work as all our models share a seed training phase, and we can subsequently presume each expert is close on the loss surface.


\section{Conclusion}
\label{sec:conclusion}
We lay groundwork for a promising new strategy whereby one can efficiently simulate principled, fine-grained ablations of training datasets and data mixtures. We are hopeful that researchers and practitioners can follow our recommendations to find beneficial data mixtures for their own models and data. Moreover, we encourage others to experiment in additional settings and report observations specific to their own assumptions and settings, particularly in relation to larger scales of models and data and limitations of the current method.

\newpage
\section*{Limitations}
\label{subsec:limitations}

Although our proposed strategies may yield immediate practical benefits certain settings, there are several limitations in the current study that call for further investigation:

\paragraph{The role of a shared optimization trajectory.} In our experiments, our models share a substantial amount of shared initial ``seed'' pre-training (42 billion tokens), and comparatively little continued pre-training ($\leq$ 27 billion tokens). Although there is precedent for multi-stage pre-training of LLMs, where greater care is taken in later stages to curate appropriate training datasets \citep{Li2022BranchTrainMergeEP, Gururangan2023ScalingEL}\footnote{Additionally, the the OLMo 1.7 \href{https://blog.allenai.org/olmo-1-7-7b-a-24-point-improvement-on-mmlu-92b43f7d269d}{blog post} describes using a 2 stage curriculum that our method may be considered an approximation of.}, there are several unanswered questions about the necessity and role of the initial optimization shared across our models -- e.g. the \textit{minimum} amount of shared initial ``seed'' pre-training necessary, or the \textit{maximum} amount of continued pre-training allowed, for models trained in parallel on partitions of a data mixture to remain ``mergeable'' and predictive of a model trained sequentially on the entire data mixture.
\paragraph{Varying data mixture proportions.} From the present study's results, one might reasonably hypothesize that a data mixture that is predicted to be particularly performant on an evaluation set(s) of interest should be upsampled if used as part of a larger training corpus. However, we leave to future work the explicit investigation of proxy metrics for predicting perplexity performance on different proportions of the same dataset components.
\paragraph{Varying total amount of training.} In our settings, we assume similar amounts of data (or logical partitions) are being mixed in each candidate configuration. Future work may investigate the viability of our method for predicting performance in lifelong learning settings, modeling the effect of adding training data to a mixture progressively.
\paragraph{Scaling up beyond 1.1b parameters.} The overall scale of our models and data is smaller than what is commonly used in practice today. Although we have promising evidence that models at the 1.1b parameter scale do not behave drastically differently from our 130m parameter models with respect to our reported trends (and that the smaller models' \textsc{merged} performance has a useful correlation with the 1.1b models' \textsc{seq} performance), we acknowledge the possibility that the patterns may be less reliable in models at 7b or larger scales, or that it may be necessary to adopt certain strategies to handle additional logistical challenges of working with larger models and corpora (e.g. larger model storage size on disk, and more copies stored for larger corpora). We are hopeful that follow-up work can allow the community to develop a stronger sense of these limitations and necessary adaptations.
\paragraph{Downstream task evaluations.} In our study, we evaluate our models with per-token perplexity score on a diverse set of  evaluation sets. The small scale of most of our models are such that useful behavior on standard downstream tasks typically requires few-shot learning or fine-tuning. Although perplexity scores from the same model can be useful for comparison and can be an informative signal about model \textit{fit} to various textual domains, we recognize that perplexity alone does not necessarily capture the aspects of language model \textit{behavior} that may be of interest, and perplexity scores by themselves are not necessarily interpretable in comparisons between domains, especially as perplexity can be highly impacted by formatting and structure of text.

\section*{Ethical Considerations}

While our work is aimed at reducing the computational requirements of experimentation with pre-training data mixtures, our experimentation itself used significant amounts of compute. We estimate that, given access to similar hardware, replicating our experiments on our small models alone would require at minimum 2,000 GPU hours (partially due to the necessity of training \textsc{seq} models for all experiments). In general, our approach should enable more frequent experimentation and testing of data mixtures due to lowered computational costs of conducting them, which may ideally enable a more nuanced understanding of the role of different types of data in different data mixtures, including with respect to fit to various evaluation domains. However, we recognized that, by Jevons Paradox, the efficiency gains introduced by our methods could lead to increased overall resource consumption rather than a reduction.


\newpage
\section*{Acknowledgments}
The authors are grateful to Luca Soldaini and Kyle Lo for assistance with procuring training data, members of the AllenNLP team for input during early stages of the project, AI2 IT for critical technical support, and the anonymous reviewers and SAC for their time and helpful feedback.
We would like to thank Joel Mire, Akhila Yerukola, Cathy Jiao, and Jared Fernandez for extremely helpful feedback.
This work was supported in part by the National Science Foundation Graduate Research Fellowship Program under grant DGE2140739.

\bibliography{anthology, custom}
\newpage
\appendix

\section{Appendix}
\subsection{S2ORC dataset details}
\label{sec:appendix_s2orc}
Tables~\ref{tab:s2orc_multiple_fos} and~\ref{tab:s2orc_sizes} contain additional information about the S2ORC text dataset. For details about exact partitioning, see our released S2ORC continued pre-training dataset at \url{https://huggingface.co/datasets/claran/modular-s2orc}. (We release this dataset as part of a collection of other datasets and models relevant to this work\footnote{\url{https://huggingface.co/collections/claran/scalable-data-ablations-6711b7078273ae67b80afc6a}}).

\begin{table}[h!]
\small
    \centering
    \begin{tabular}{c|r|c}
        \# FoS & Document Count & \% \\
        \hline
        0 & 210 & $1.0\%$\\
        1 & 15844 & $76.8\%$\\
        2 & 4318 & $20.9\%$ \\
        3 & 259 & $1.3\%$\\
        4 & 7 & $<0.1\%$\\
        5 & 1 & $<0.1\%$\\
        \hline
        \textit{All (sample)} & 20,639 & $100\%$ \\
    \end{tabular}
    \caption{Distribution of number of tagged fields of study in S2ORC documents, the ~20k documents in our validation set. The original S2ORC corpus contains 81.1M documents in total. We use only the first field of study for our partitioning purposes.}
    \label{tab:s2orc_multiple_fos}
\end{table}

\begin{table*}[h!]
\small
    \centering
    \begin{tabular}{c|c||c|c}
        Field of Study & \# Partitions & Field of Study & \# Partitions \\
        \hline
        Agricultural and Food Sciences & 5 &        History & 1 \\
        Art & 1 &                       Linguistics & 1 \\
        Biology & 18 &                  Materials Science & 5 \\
        Business & 4 &                  Mathematics & 12 \\
        Chemistry & 4 &                 Medicine & 14 \\
        Computer Science & 11 &         NA & 2 \\
        Economics & 4 &                 Philosophy & 1 \\
        Education & 3 &                 Physics & 20 \\
        Engineering & 5 &               Politics & 2 \\
        Environmental Science & 7 &     Psychology & 6 \\
        Geology & 1 &                   Sociology & 1 \\
        \hline
        \textbf{Total} & \textbf{128} \\

    \end{tabular}
    \caption{We partition S2ORC topically and temporally. There are 128 partitions total, with a median token count of 287 million tokens. Partition sizes vary no more than about one order of magnitude in token count.}
    \label{tab:s2orc_sizes}
\end{table*}

\subsection{Licenses}
S2ORC and M2D2 have CC BY-NC licenses. Out of the other Paloma subsets we used, most are licensed under AI2 ImpACT License - Low Risk Artifacts, excepting Wikitext-103 (CC BY-SA) and RedPajama. Our use of the datasets is for research purposes and aligns with their intended uses.

\subsection{Training details}
\label{sec:appendix_hyperparams}

For all models, we preserve optimizer states between seed and continued pre-training, use an Adam optimizer with 0.1 weight decay and (0.9, 0.95) betas, and train on batch sizes of 1024 comprised of packed sequences of 2048 tokens, or 2M tokens per batch.

In seed pre-training, we use a cosine learning rate with warmup up to a maximum of $6e-4$ followed by annealing to $6e-5$ by the end of seed training. However, we find it beneficial to jump back up to the maximum LR when starting from a seed model initialization and switching to optimization on the domain-specific partitions.

All experiments use at most a single node at a time. We used between 1 GPU (modular training) and 8 GPUs (longest \textsc{seq} training) per model training job. The GPU hardware available to us was a mixture of A6000s, L40s, and 80GB A100s.

\subsubsection{A note on model architecture}

We describe our model as ``PaLM-like'' to distinguish between sequential and parallel use of attention and MLP blocks: Llama models have sequential blocks, while PaLM and our models have parallel blocks. We do not believe that the use of sequential vs parallel MLP and attention blocks would fundamentally affect our results or imply any loss of generality. On the other hand, one benefit to using a parallel architecture was that we observed superior throughputs on our setup early on, and this allowed us to run more data mixture experiments with our small models.

\subsection{Modeling entire fields of study}
\label{sec:appendix_s2orc_fos}

\begin{figure*}[h]
    \centering
    \includegraphics[width=\textwidth]{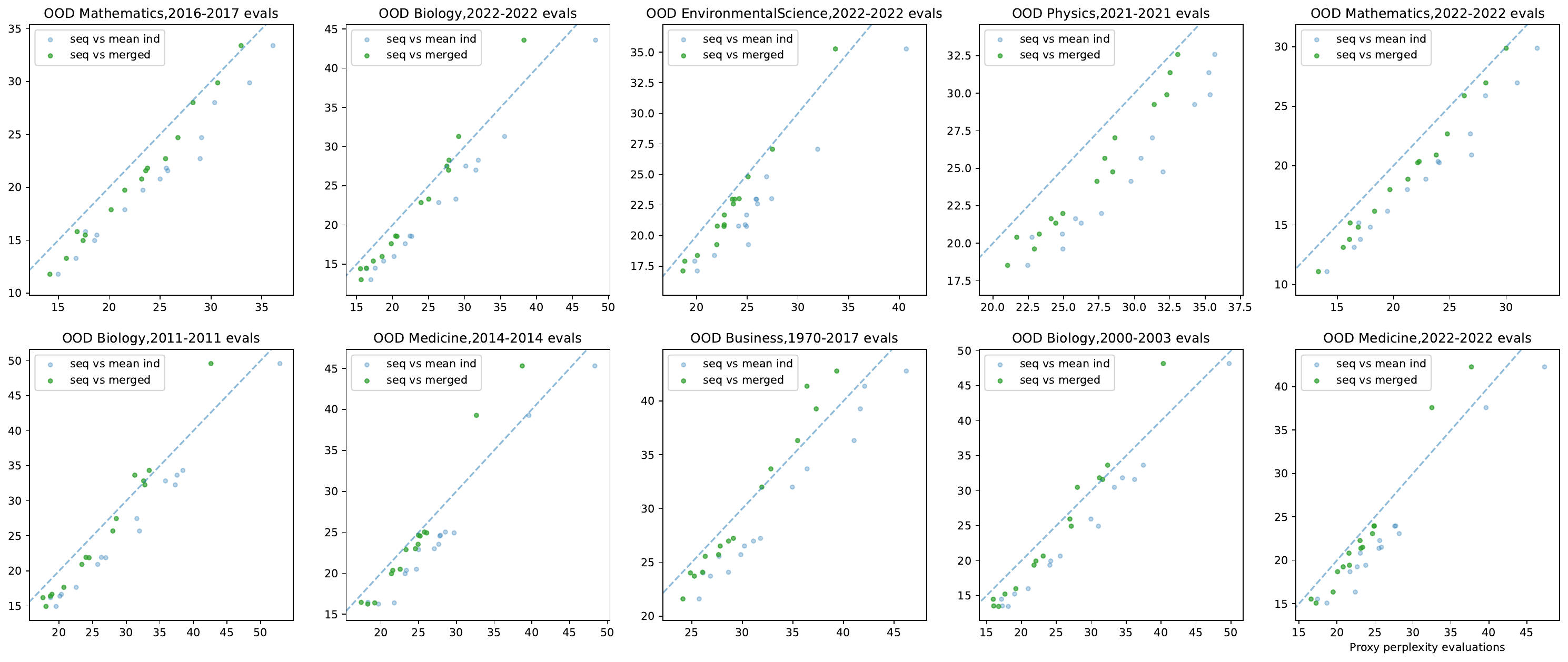}
    \caption{Here, we consider entire S2ORC fields of study as their own data mixtures. We display ten randomly selected S2ORC domains' evaluation scores (each scatterplot corresponds to one evaluation domain). \textsc{seq} models are trained on the entire field of study, and \textsc{merged} models are the parameter average of multiple models, as few as 2 or as many as 20. \textsc{ind} scores are the macro-average of the individual models' scores on the evaluation domain. Each point in a figure corresponds to the \textsc{merged} or \textsc{ind} score plotted against the \textsc{seq} score for a single field of study. As before, the y-axis shows \textsc{seq} perplexity evaluation performance, while the x-axis shows proxy perplexity scores.}
    \label{fig:s2orc_byfos_ood_evals}
\end{figure*}

We select partitions $\mathcal{P}$, each of which contains between 1 and 20 ``base unit'' partitions $\mathcal{P'}$. Note that here, there is no overlap between different candidate data mixtures. Instead, we present this as a transition between the earlier settings mixing evenly sized $\mathcal{P'}$ and the later settings mixing unevenly sized $\mathcal{P}$. We focus on 1) providing further evidence of the ID vs OOD behavior seen in the previous experiments, and 2) providing a basis for the ``macro''-\textsc{merged} model evaluation strategy we see in experiments testing uneven data mixtures.

In Figure~\ref{fig:s2orc_byfos_ood_evals} we see that \textsc{seq} scores are indeed strongly correlated with the performance of the aggregated \textsc{merged} model for each OOD field of study.

\subsection{Fixing one partition}
\label{sec:appendix_fixwiki}

\begin{figure}[h]
    \centering
    \begin{subfigure}[b]{0.48\textwidth}
        \includegraphics[width=\textwidth]{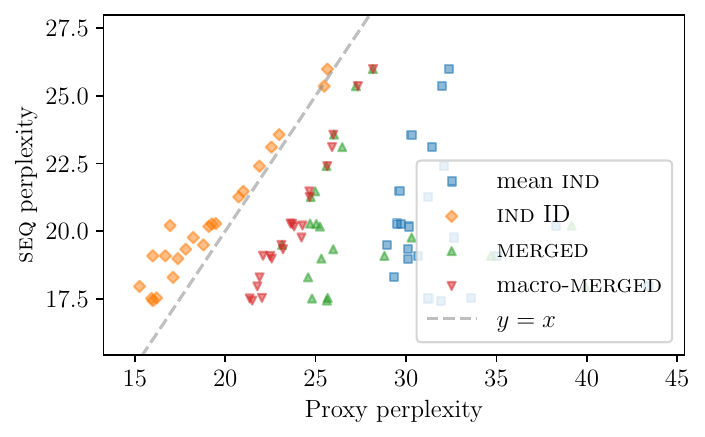}
        \label{fig:fos_fixL1_combo_in_domain}
    \end{subfigure}
    \hfill
    \begin{subfigure}[b]{0.48\textwidth}
        \includegraphics[width=\textwidth]{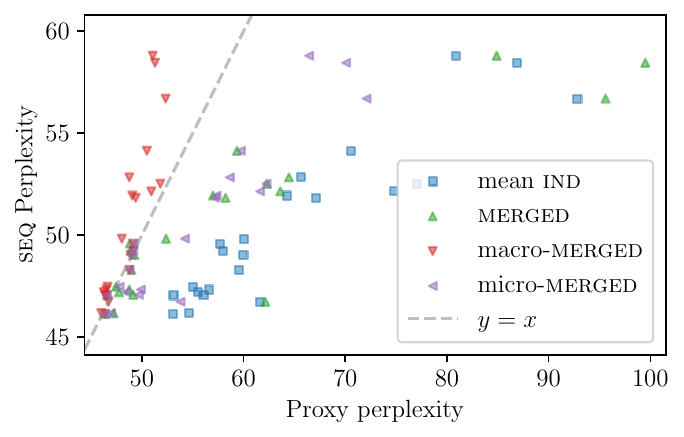}
        \label{fig:fos_fixL1_combo_macroavg}
    \end{subfigure}
    \caption{\textsc{seq} vs. proxy macro-averaged Paloma perplexity scores of models trained on data mixtures where all contain the same L1 Wikipedia domain. It is again beneficial to use \textit{micro-}\textsc{merged} models: 0.935 corr. with \textsc{seq}, vs. 0.889 (\textsc{merge} of the \textsc{seq} models for each $\mathcal{P}$), 0.920 (mean \textsc{ind} scores), 0.882 (macro-\textsc{merged}), or 0.794 (mean \textsc{merged} scores)}.
    \label{fig:fos_fixL1_combo}
\end{figure}

This experiment is a practical extension and variation of the experiment in \S\ref{subsec:fos_l1_mix}. Figure~\ref{fig:fos_fixL1_combo} contains the results of this study. We again see that micro-\textsc{merged} model scores are the most useful proxy score overall. For 3 out of 9 Paloma splits, selecting the ``best'' mixture by proxy metric gives us the ``best'' model as measured by evaluating full \textsc{seq} model for the data mixture on the same evaluation set, and in another 3, the ``best'' mixture appears in the top 3 mixtures chosen by the proxy ranking (out of 22 candidates). Selecting the top candidate by proxy gives us a median ``true'' rank of 2, with a maximum ``true'' rank of 5, across the 9 Paloma splits we evaluate on.

\subsection{Larger models}
\label{sec:appendix_1b}

\begin{figure}[h]
    \centering
    \begin{subfigure}[b]{0.45\textwidth}
        \includegraphics[width=\textwidth]{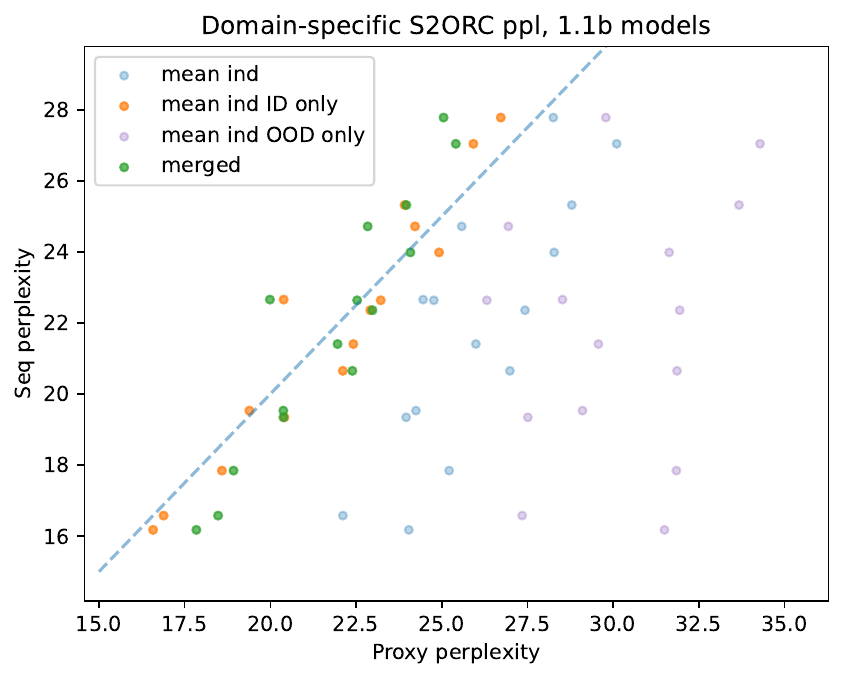}
        \subcaption{Here, the means of the in-domain scores of \textsc{ind} models is not as clearly better than the \textsc{merged} models for predicting \textsc{seq} performance on data mixtures, but the correlation is still strong.}
        \label{subfig:1bil_in_domain}
    \end{subfigure}
    \hfill
    \begin{subfigure}[b]{0.45\textwidth}
        \includegraphics[width=\textwidth]{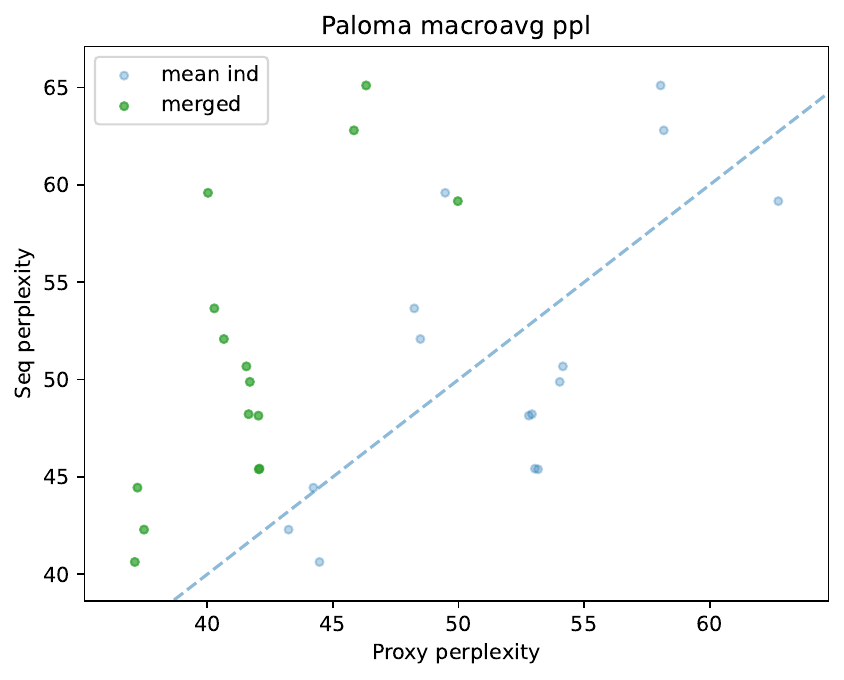}
        \subcaption{On OOD Paloma evals the correlation is even weaker, but we still see that we can pick the ``best'' data mixtures by picking the ``best'' model mixtures.}
        \label{subfig:1bil_macroavg}
    \end{subfigure}
    \caption{At larger 1.1b model scale, the correlations we see are weaker but still usable in practice if one only wishes to select the most performant data mixtures. Some of the increased variance seems to come from the sample itself (just the smallest fields of study $\mathcal{P}$ from S2ORC). We see a similarly increased amount of variance in the 130m models when replicating on the same data mixtures (see Figure~\ref{subsec:appendix_130_repl}).}
    \label{fig:s2orc_1bil_rand2}
\end{figure}

Figure~\ref{fig:s2orc_1bil_rand2} plots \textsc{seq} vs. proxy metrics in 1.1b parameter models. We study all $\{\vec{d} \subseteq \{0,1\} : ||\vec{d}_{S2ORC}||_1=2, ||\vec{d}_{Wiki}||_1=0, |\mathcal{P}| = 1 ~\forall~ \mathcal{P} \in \vec{d}\}$.

Though the correlation values themselves are weaker, this is also true in the replication of these experiments with smaller models as both \textsc{seq} and proxy models, seen in Appendix~\ref{subsec:appendix_130_repl}. Moreover, in all settings and evaluation domains, the data mixture with the lowest proxy perplexity score also leads to one of the lowest perplexity scores in \textsc{seq} models. 

\subsubsection{Predicting larger model performance}
\label{subsec:appendix_1b_pred}

Here, we use these same candidate mixtures to present another version of the experiment shown in \S\ref{subsec:1b_expts}, where we explore the possibility of predicting \textsc{seq} model performance of larger models using smaller, proxy models. Figure~\ref{fig:s2orc_1bil_pred} depicts the results of this experiment.

\begin{figure}[h]
    \centering
    \begin{subfigure}[b]{0.45\textwidth}
        \includegraphics[width=\textwidth]{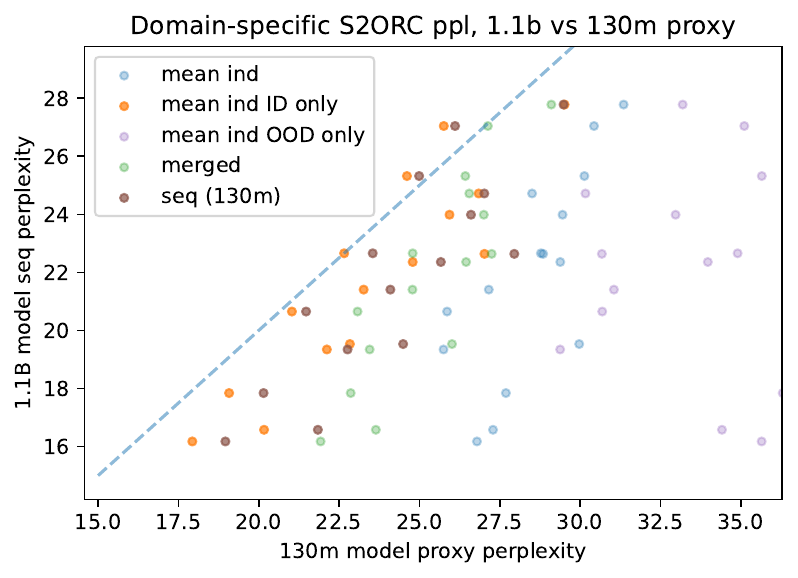}
        \subcaption{These are the weakest correlation so far for predicting in-domain \textsc{seq} performance, but still a recognizable trend. The weaker correlation is unsurprising given that we are comparing models different in scale by a factor of almost 10.}
        \label{subfig:1bil_pred_in_domain}
    \end{subfigure}
    \hfill
    \begin{subfigure}[b]{0.45\textwidth}
        \includegraphics[width=\textwidth]{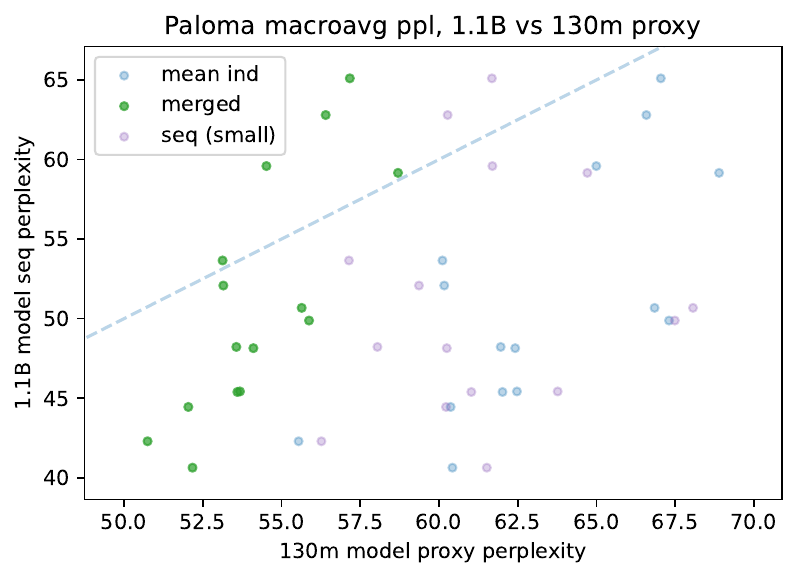}
        \subcaption{Interestingly, scores of the smaller \textsc{merged} models are actually much \textit{more strongly correlated} with \textsc{seq} model scores than scores of smaller \textsc{seq} models!}
        \label{subfig:1bil_pred_macroavg}
    \end{subfigure}
    \caption{Concurrent work by \citet{ye2024data} shows that smaller (\textsc{seq}) models can predict larger (\textsc{seq}) model performance on a data mixture. While we focus on asymptotic efficiency gains from reusing training performed on decomposed subparts of an arbitrary data mixture, we show that these efficiency gains are \textit{compatible}.}
    \label{fig:s2orc_1bil_pred}
\end{figure}

\subsubsection{Replication in 130m parameter models}
\label{subsec:appendix_130_repl}

We replicate the experiments from \S\ref{sec:appendix_1b} and \S\ref{subsec:appendix_1b_pred} using solely 130m parameter models. We observe that this set of data mixtures yields noisy correlations compared to the other settings we explore at this smaller scale. We conjecture that this may be in part due to the types of fields of study that are small enough to have only one base component partition: Art, Geology, History, Linguistics, Philosophy, and Sociology. These are mostly humanities fields with high overlap and similarity as measured by pairwise cosine distances between mean SentenceBERT \citep{reimers-gurevych-2019-sentence} embeddings of held-out examples from each of these partitions. Specifically, we observe separately that relaxing the definition of a data mixture $\vec{d}$ to allow for multiple copies of the same partition in a single mixture can cause the associated mixture to deviate significantly from the other points in the study -- accordingly, it could be the case that partitions that are too similar to each other (such that, for example, a human might have trouble distinguishing between them as separate domains) may behave differently in a data mixture compared to other, less similar partitions.

\begin{figure}[h]
    \centering
    \begin{subfigure}[b]{0.45\textwidth}
        \includegraphics[width=\textwidth]{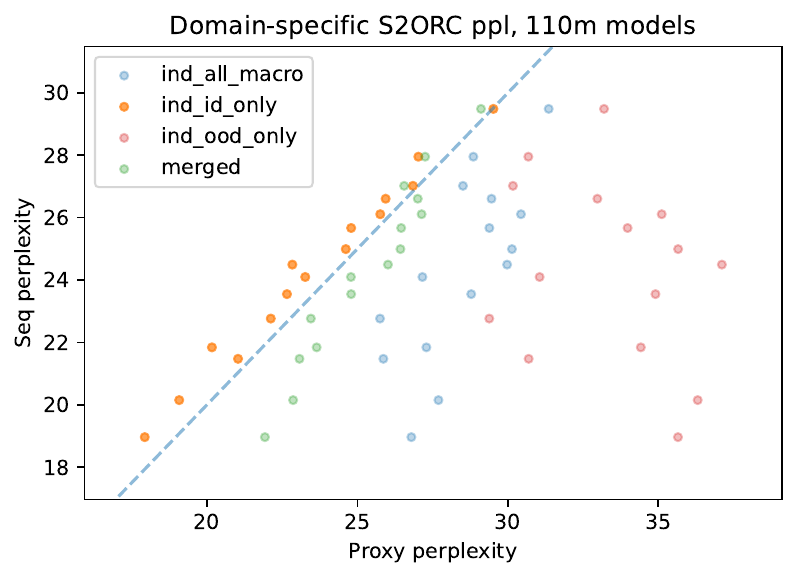}
        \label{subfig:1bil_pred_in_domain}
    \end{subfigure}
    \hfill
    \begin{subfigure}[b]{0.45\textwidth}
        \includegraphics[width=\textwidth]{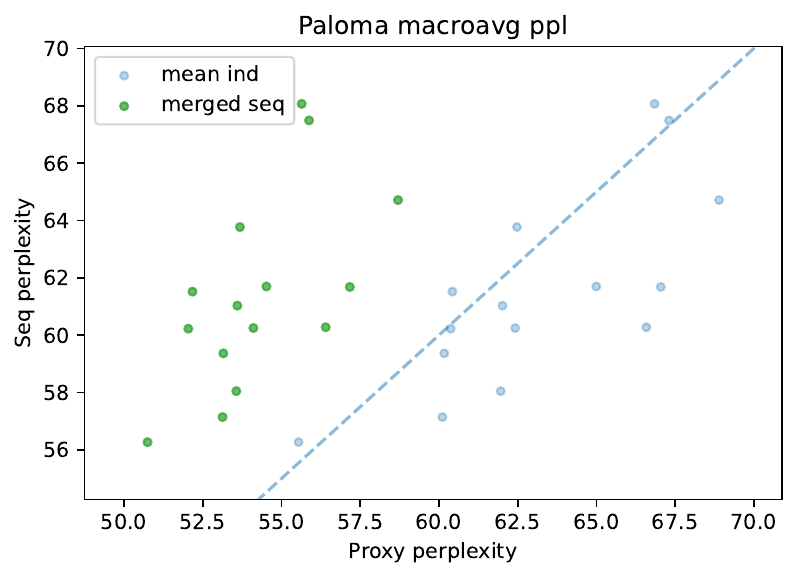}
        \label{subfig:1bil_pred_macroavg}
    \end{subfigure}
    \caption{We see that the weaker correlations seen in Figures~\ref{fig:s2orc_1bil_rand2} and \ref{fig:s2orc_1bil_pred} seem unlikely to be solely due to the larger model scale, as the correlation is weaker also when 130m models are used as both \textsc{seq} and proxy models.}
    \label{fig:s2orc_1bil_pred}
\end{figure}

\end{document}